\documentclass[10pt,letterpaper,twocolumn]{article}
\usepackage[T1]{fontenc}
\usepackage{newtxtext}
\usepackage{amsmath}
\usepackage{graphicx}
\usepackage[margin=0.75in,columnsep=0.25in]{geometry}
\usepackage{natbib}
\usepackage[hyphens]{url}
\usepackage[hidelinks]{hyperref}

\bibpunct{(}{)}{;}{a}{}{,}
\let\cite\citep
\title{GaokaoMathGen: Chinese Gaokao Mathematics Finale Problem Generation Controlled by Latent Reasoning Structures}
\author{%
  \begin{minipage}{0.97\textwidth}
    \centering
    \begin{tabular*}{\linewidth}{@{}c@{\extracolsep{\fill}}ccc@{}}
      Qian Chen\textsuperscript{1,}\thanks{Qian Chen and Kun Zhu contributed equally to this work.}
      & Kun Zhu\textsuperscript{2,*}
      & Zhongyi Wang\textsuperscript{3}
      & Ling Chen\textsuperscript{1}\\[0.2em]
      \small\href{mailto:qianchencs@cs.zju.edu.cn}{qianchencs@cs.zju.edu.cn}
      & \small\href{mailto:kunzhu@zstu.edu.cn}{kunzhu@zstu.edu.cn}
      & \small\href{mailto:yzy5630@mail.ustc.edu.cn}{yzy5630@mail.ustc.edu.cn}
      & \small\href{mailto:lingchen@cs.zju.edu.cn}{lingchen@cs.zju.edu.cn}
    \end{tabular*}
    \endgraf\medskip
    {\small
      \textsuperscript{1}State Key Laboratory of Blockchain and Data Security,\\
      College of Computer Science and Technology, Zhejiang University\\[0.2em]
      \textsuperscript{2}Zhejiang Sci-Tech University\qquad
      \textsuperscript{3}iflytek\endgraf
    }
  \end{minipage}%
}
\hypersetup{pdfauthor={Qian Chen, Kun Zhu, Zhongyi Wang, Ling Chen}}

\hypersetup{
  pdftitle={GaokaoMathGen: Chinese Gaokao Mathematics Finale Problem Generation Controlled by Latent Reasoning Structures},
  pdfkeywords={Mathematical problem generation, Chinese Gaokao, latent reasoning structures, controllable generation, difficulty diagnostics}
}
\date{}

\begin{document}

\maketitle

\begin{abstract}
Generating Chinese Gaokao mathematics finale problems requires controlling not only the included skills, but also the underlying mathematical kernel and solution path. Existing LLM-based methods mainly rely on surface-level specifications, while the realized difficulty of generated problems cannot be determined directly from prompted targets. Therefore, we propose GaokaoMathGen, a framework for Chinese Gaokao mathematics finale problem generation controlled by latent reasoning structures. It extracts standardized latent reasoning structures from seed problems and performs constrained synthesis using skills, methods, a core identity, and a pivot step as structural controls. We further estimate realized difficulty through logical-quality filtering, fixed-solver decoding effort, structural features, and solver-output search signals. Experiments show that semantic normalization improves cross-annotator comparability and that 78.0\% of 609 scored candidates pass the logical-quality filter. In a teacher audit of the 60 highest-effort candidates, all were judged to reach finale or penultimate problem difficulty, and only one exhibited a severe logical issue. These results support latent reasoning structures as effective control signals for generating high-difficulty mathematical problems.
\end{abstract}

\smallskip
\begingroup\small\raggedright
\par
\endgroup

\section{Introduction}

The Chinese National College Entrance Examination, known as Gaokao, is one of the most influential high-stakes examinations in China \cite{jia2025highest}. Its mathematics finale problems are designed to be highly discriminative: they often require students to integrate multiple knowledge points, identify hidden mathematical structure, perform non-trivial transformations, and organize a multi-step solution path. Accurate and controllable generation of such problems is valuable for scalable reasoning-oriented training, targeted student practice, and more reliable evaluation of mathematical reasoning models. This task remains challenging, as a generated problem must be mathematically valid, aligned with the senior-high-school curriculum, solvable through a coherent reasoning path, and calibrated to an intended difficulty level.

Mathematical problem generation has been studied through expert-authored construction, rule- or template-based generation \cite{xu2021procedural,yumetamath}, and recent LLM-based methods \cite{gunasekar2023textbooks,lu2024mathgenie,shah2407ai}. Expert-authored problems provide strong guarantees on validity and educational value, but they require substantial human expertise and are difficult to scale. Rule- and template-based methods improve controllability by instantiating predefined item models or solution patterns, but they are less suited to generating diverse non-routine finale problems. LLM-based methods offer better scalability and linguistic diversity, and recent work has improved difficulty by extracting topics or skills from existing datasets and recombining them during generation \cite{tangmathscale}. For example, MATH$^2$ \cite{shah2407ai} constructs difficult problems by combining core skills extracted from the MATH dataset. These studies show the promise of LLM-assisted mathematical problem generation. In most existing pipelines, the control signal is specified at the level of topics, skills, templates, equations, or seed-problem transformations.

However, for Chinese Gaokao mathematics finale problems, such surface-level control is insufficient. The difficulty of a finale problem is not determined only by which knowledge points appear, but also by how they are grounded in a mathematical kernel, concealed in the problem statement, and recovered through a solution bottleneck. Two problems may involve the same skills and methods, yet differ substantially in difficulty depending on whether the key identity, auxiliary construction, or transformation step is explicit, implicit, or only recoverable after several intermediate deductions. This observation motivates a shift from surface-level control to structure-level control: the generator should specify not only the intended skills and methods, but also the underlying mathematical core and the key pivot step that make the problem solvable and discriminative.

Therefore, we study the Chinese Gaokao mathematics finale problem generation controlled by this structural view. The task presents two main challenges. First, \textbf{the mapping from latent reasoning structures to well-formed problems is underdetermined}. A latent reasoning structure specifies the intended skills, methods, core identity, and pivot step, but not how they should be encoded as conditions and subproblems that form a valid reasoning path and resist elementary shortcuts. Direct LLM generation may consequently realize these elements only superficially, producing shortcut solutions, redundant conditions, or invalid mathematical cores. Second, \textbf{realized difficulty is not directly observable from the generation target}. A requested difficulty level expresses the intended difficulty but does not demonstrate the reasoning effort required by the generated problem. Teacher annotation is reliable but costly, while standalone LLM judgments can exhibit systematic biases and solver accuracy provides limited diagnostic evidence \cite{zheng2023judging}.

To address these challenges, we propose GaokaoMathGen, a framework for Chinese Gaokao mathematics finale problem generation controlled by latent reasoning structures. To the best of our knowledge, this is the first study dedicated to Chinese Gaokao mathematics finale problem generation. Moving beyond surface-level content specification, GaokaoMathGen organizes standardized skills and methods around a core identity and pivot step, which serve as structural controls for candidate synthesis. The main contributions are summarized as follows:

\begin{itemize}
\item We propose a two-stage generation framework in GaokaoMathGen for Chinese Gaokao mathematics finale problems. The first stage converts seed finale problems into standardized latent reasoning structures. The second stage performs constrained synthesis controlled by these latent reasoning structures. Standardized skills and methods specify the required reasoning, while the core identity and pivot step anchor the concealed mathematical kernel and coherent solution path.

\item We develop a low-cost realized-difficulty estimation protocol based on core-anchored structural signals, fixed-solver decoding effort, and teacher audit. This protocol avoids relying solely on prompted difficulty labels and provides diagnostic evidence about whether generated problems realize the intended reasoning difficulty.

\item We conduct systematic experiments on seed labeling, problem construction, quality evaluation, and ablation settings. The results show that core-anchored construction improves structural controllability and produces high-effort candidate problems.
\end{itemize}

\section{Related Work}

\subsection{Mathematical Problem Generation}

Mathematical problem generation has been studied through expert-authored construction, rule- or template-based generation, and LLM-based generation. Expert-authored construction, including human-written examination problems and expert-designed item models, can provide strong guarantees on mathematical validity, curriculum alignment, and educational value, but it requires substantial human expertise and is difficult to scale \cite{shah2407ai}. Rule- and template-based methods improve scalability by instantiating predefined item models, equations, or procedural templates \cite{xu2021procedural}. These methods offer controllable formats and solution patterns, but their reliance on explicit templates limits their ability to generate diverse non-routine problems.

Recent LLM-based methods further improve scalability and linguistic diversity. Prior work generates mathematical word problems from equations, topics, or contexts while enforcing mathematical consistency or difficulty-related constraints \cite{wang2021math,jiao2023automatic}. For example, MATHWELL \cite{christ2024mathwell} uses teacher annotations to improve the solvability, accuracy, and educational appropriateness of generated math word problems. MathGenie \cite{lu2024mathgenie} synthesizes new math problems through solution augmentation and question back-translation. Closely related to difficult problem generation, MATH$^2$ \cite{shah2407ai} extracts core skills from the MATH dataset and prompts LLMs to generate problems combining pairs of skills.

This line of work demonstrates the promise of scalable mathematical problem generation, but its control signals are usually specified at the level of topics, equations, skills, templates, or seed-problem transformations. Such surface-level control is not sufficient for Chinese Gaokao mathematics finale problems, where difficulty depends not only on what knowledge points are involved, but also on how a mathematical kernel is concealed, transformed, and recovered through a coherent reasoning path. 

In contrast, our proposed framework, GaokaoMathGen, focuses on structure-level control by representing each intended generation target as a latent reasoning structure comprising skills, methods, a core identity, a pivot step, and a target difficulty requirement.

\subsection{Reasoning Evaluation and Difficulty Estimation}

Mathematical reasoning benchmarks provide the standard setting for measuring solver performance on multi-step problems. Datasets, e.g., GSM8K \cite{cobbe2021trainingverifierssolvemath} and MATH \cite{hendrycks2021measuring}, evaluate grade-school and competition-style reasoning, while recent benchmarks extend evaluation to high-stakes examinations, olympiad-level problems, Chinese mathematical tasks, and more fine-grained difficulty annotations \cite{zhang2023evaluating,he2024olympiadbench,zhong2024agieval,liu2024mathbench,liu2025finemath}. These benchmarks are valuable for comparing reasoning models, but their difficulty information is usually attached to human-authored problems through source labels, expert annotation, or aggregate model performance. As a result, they offer limited guidance for determining whether a newly generated problem realizes a specified reasoning structure.

Difficulty control and estimation have also been studied in question generation and educational assessment. Prior methods often rely on prompted difficulty levels, surface features of the problem statement, equation or solution length, expert labels, or solver correctness \cite{cheng2021guiding,jiao2023automatic}. These signals are useful for scalable assessment, but they are not sufficiently diagnostic for Gaokao mathematics finale problems. In this setting, a problem can appear difficult for undesirable reasons, e.g., ambiguity or excessive computation, while the intended difficulty should come from how skills and methods are organized around a hidden mathematical kernel and recovered through a solution bottleneck.

Our proposed framework, GaokaoMathGen, addresses this gap by treating realized difficulty as a diagnostic property of generated candidates rather than as a prompted difficulty label. It estimates difficulty from lightweight post-generation evidence, including logical quality, fixed-solver effort, skill-method structural signals, and solver-output search proxies. This connects difficulty estimation to structure-level control while avoiding direct reliance on prompted difficulty labels or final solver correctness.

\section{Problem Formulation}

We study controllable generation of Chinese Gaokao mathematics finale problems from a structure-level perspective. Given a seed mathematics problem \(x_i\) and a target difficulty \(\tau_i\), our goal is not to imitate the surface form of \(x_i\), but to extract its reusable reasoning information and synthesize a new candidate problem under the specified difficulty.

For a seed problem \(x_i\), we construct its intended reasoning structure \(z_i\) as follows:
\begin{equation}
\begin{aligned}
z_i &= \Psi(x_i,\tau_i) \\
    &= (\mathcal{S}_i,\mathcal{M}_i,c_i,p_i,\tau_i),
\end{aligned}
\label{eq:latent-structure}
\end{equation}
where \(\Psi\) denotes the structure extraction and assembly process, \(\mathcal{S}_i\subseteq\mathcal{V}^S\) is a finite set of standardized skills, and \(\mathcal{M}_i\subseteq\mathcal{V}^M\) is a finite set of standardized methods. The core identity \(c_i\) and pivot step \(p_i\) are variable-length textual descriptors rather than fixed-dimensional vectors or labels selected from a closed vocabulary. The target difficulty \(\tau_i\) is an ordinal value from the target-difficulty scale used during synthesis. Accordingly, \(z_i\) is a heterogeneous structured tuple rather than a fixed-dimensional representation.

To reduce annotator-specific variation, raw skill and method labels produced by multiple LLM-based annotators are semantically normalized into shared vocabularies. Standardized skills refer to canonical labels for the mathematical knowledge required by a problem, e.g., derivative monotonicity, conic parameterization, or recurrence transformation. Standardized methods refer to canonical labels for the reasoning strategies used to combine these skills, e.g., substitution, auxiliary construction, or case analysis. The core identity denotes the mathematical kernel supporting the solution, e.g., a theorem, extremal property, or structural equivalence. The pivot step denotes the key transformation through which a solver connects the problem statement to this kernel.

Given \(z_i\), the generation goal is written as follows:
\begin{equation}
\hat{x}_i=G(z_i),
\label{eq:generation-goal}
\end{equation}
where \(\hat{x}_i\) is a candidate finale problem generated by GaokaoMathGen. The requirement is not only that \(\hat{x}_i\) contains the specified skills and methods, but also that its solution path coordinates them around the core identity and pivot step under the target difficulty. This representation is intentionally different from a chain-of-thought trace \cite{wei2022chain}: it describes the intended reasoning structure that a problem should realize, rather than an observed solver trajectory.

\section{Method}

\subsection{Overview}

\begin{figure*}[t]
    \centering
    \includegraphics[width=\textwidth]{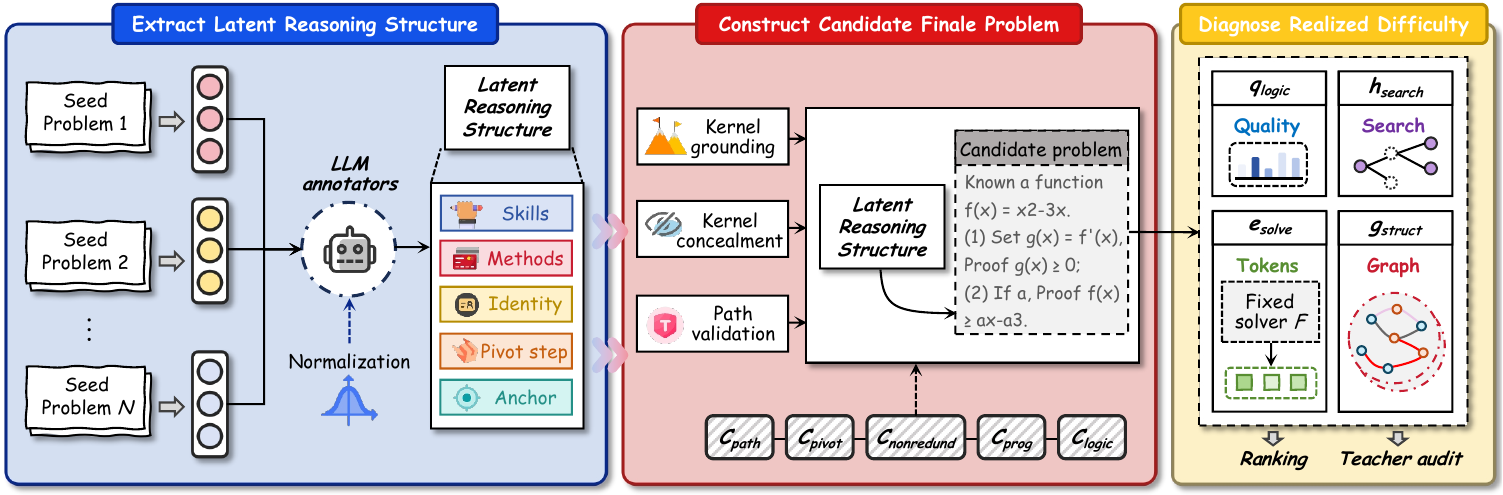}
    \caption{Overview of GaokaoMathGen. The framework extracts standardized latent reasoning structures from seed problems, performs structure-guided synthesis under explicit constraints, and collects post-generation diagnostic evidence for realized difficulty.}
    \label{fig:framework}
\end{figure*}

As illustrated in Figure~\ref{fig:framework}, GaokaoMathGen treats controllable generation as a process of extracting, realizing, and diagnosing latent reasoning structures. It first extracts and standardizes latent reasoning structures from seed mathematics problems, forming a collection of structure-level generation targets. It then synthesizes candidate finale problems under structural constraints, so that the generated statements are guided by the intended core identity and pivot step rather than by surface topic or difficulty descriptions alone. Finally, it collects diagnostic evidence for realized difficulty, including logical quality, fixed-solver decoding effort, skill-method structural signals, and search-related signals extracted from solver outputs. These stages respectively correspond to obtaining the reasoning specification, realizing it as a well-formed candidate problem, and estimating whether the generated problem exhibits evidence of the intended difficulty.

\subsection{Reasoning Structure Extraction}

Raw seed problems provide the source material for generating Chinese Gaokao mathematics finale problems. However, they cannot be used directly as structure-level control signals, as their reusable reasoning information is entangled with surface wording, problem-specific notation, and annotator-dependent descriptions. To obtain such control signals, GaokaoMathGen extracts a standardized latent reasoning structure from each seed problem, allowing later generation to be conditioned on comparable structural specifications rather than on the original problem statement.

For each seed problem \(x_i\), multiple LLM-based annotators produce structured annotations that identify the problem's skills, methods, mathematical kernel, difficulty anchor, and auxiliary metadata. The annotated skills and methods provide raw label sets \(R_i^S\) and \(R_i^M\), while the mathematical-kernel fields provide the core identity \(c_i\) and pivot step \(p_i\). The difficulty annotation provides a seed-level difficulty anchor \(\delta_i\), which contextualizes the target difficulty \(\tau_i\) specified for synthesis but is not itself treated as the generation target. Subject-area metadata further constrains skill normalization, reducing the risk of merging labels that are lexically similar but mathematically different across domains. Implementation details of the annotation and normalization protocol are provided in the Technical Supplement.

To reduce variation across LLM-based annotators, GaokaoMathGen semantically normalizes raw skill and method labels into shared vocabularies. The normalization mappings are defined as follows:
\begin{equation}
\pi_S:\mathcal{R}^S\rightarrow\mathcal{V}^S,
\qquad
\pi_M:\mathcal{R}^M\rightarrow\mathcal{V}^M,
\label{eq:label-normalization}
\end{equation}
where \(\mathcal{R}^S\) and \(\mathcal{R}^M\) denote raw skill and method label spaces, and \(\mathcal{V}^S\) and \(\mathcal{V}^M\) denote standardized vocabularies. Skill labels are normalized within subject areas to reduce cross-domain merging, while method labels are normalized globally, as they describe reusable reasoning strategies across mathematical domains.

After normalization, the standardized labels and mathematical-kernel annotations are combined with the target difficulty \(\tau_i\). The resulting latent reasoning structure is defined as follows:
\begin{equation}
\begin{aligned}
\mathcal{S}_i &= \{\pi_S(s)\mid s\in R_i^S\},\\
\mathcal{M}_i &= \{\pi_M(m)\mid m\in R_i^M\},
\end{aligned}
\label{eq:standardized-structure}
\end{equation}
where \(R_i^S\subseteq\mathcal{R}^S\) and \(R_i^M\subseteq\mathcal{R}^M\) are the raw skill and method sets extracted for \(x_i\), and \(\tau_i\) is the target difficulty specified for the corresponding synthesis instance.

\subsection{Constrained Structure-Guided Synthesis}

The structure collection \(\mathcal{Z}\) provides structure-level control signals, but converting these signals into well-formed finale problems remains nontrivial. A generator may include the requested skills and methods while still producing shallow testing targets, shortcut solutions, redundant conditions, loose subproblem dependencies, or invalid mathematical cores. Therefore, GaokaoMathGen performs constrained structure-guided synthesis by using the core identity and pivot step as structural anchors. The goal is to construct an exam-style statement in which the mathematical kernel is recoverable through the intended pivot step, but not directly exposed by the surface form.

For a latent reasoning structure \(z_i\in\mathcal{Z}\), the constrained synthesis process is written as follows:
\begin{equation}
\hat{x}_i=G(z_i;\mathcal{C}),
\label{eq:constrained-generation}
\end{equation}
where \(\hat{x}_i\) is the generated candidate problem and \(\mathcal{C}\) denotes the structural requirements that guide how \(z_i\) is realized as both a problem statement and an intended solution path. The core identity \(c_i\) provides the mathematical basis of the intended solution, while the pivot step \(p_i\) specifies the bottleneck through which this basis should be reached.

During synthesis, GaokaoMathGen operationalizes \(\mathcal{C}\) through three connected construction operations. \emph{Kernel grounding} organizes the standardized skills and methods around \(c_i\), so that they serve the same mathematical kernel rather than appearing as independent topical requirements. \emph{Kernel concealment} transforms the explicit kernel into a solvable but nontrivial exam-style statement, e.g., by introducing a new object, redistributing conditions, or expressing key information through constraints. \emph{Path validation} examines whether the intended solution still needs to pass through \(p_i\) after common shortcuts are considered.

To make these operations explicit and auditable, we express the structural requirements as follows:
\begin{equation}
\mathcal{C}=
\{\mathcal{C}_{\mathrm{path}},
\mathcal{C}_{\mathrm{pivot}},
\mathcal{C}_{\mathrm{nonredund}},
\mathcal{C}_{\mathrm{prog}},
\mathcal{C}_{\mathrm{logic}}\},
\label{eq:constraint-set}
\end{equation}
where \(\mathcal{C}_{\mathrm{path}}\) specifies the expected reasoning connection between problem conditions and the core identity, and \(\mathcal{C}_{\mathrm{pivot}}\) specifies that common elementary shortcuts should not bypass the pivot step. \(\mathcal{C}_{\mathrm{nonredund}}\) specifies that conditions should contribute to solvability rather than only increasing surface complexity. For multi-part problems, \(\mathcal{C}_{\mathrm{prog}}\) specifies coherent progression among subproblems. \(\mathcal{C}_{\mathrm{logic}}\) specifies mathematical consistency, clear variable ranges, valid boundary cases, and solvability. Implementation details of the synthesis protocol are provided in the Technical Supplement.

The generator's own blueprint is not reliable evidence that the structural requirements in \(\mathcal{C}\) are actually satisfied. Therefore, GaokaoMathGen applies a separate structural consistency checker. To reduce anchoring to the generator's self-explanation, the checker reviews the candidate statement together with the associated core identity and pivot step. The checker returns constraint-level diagnostics as follows:
\begin{equation}
\begin{aligned}
V(\hat{x}_i,z_i)
&=(v_{\mathrm{path}},v_{\mathrm{pivot}},v_{\mathrm{nonredund}},v_{\mathrm{prog}},v_{\mathrm{logic}})\\
&\in\{0,1\}^{5},
\end{aligned}
\label{eq:checker-output}
\end{equation}
where each component indicates whether the corresponding structural requirement in \(\mathcal{C}\) is satisfied. A candidate receives a structural pass only when all components equal 1. If any violation is detected, an update operator revises the candidate statement as follows:
\begin{equation}
\hat{x}_i^{(t+1)}
=
\Phi(\hat{x}_i^{(t)},z_i,V(\hat{x}_i^{(t)},z_i)).
\label{eq:revision-update}
\end{equation}

This checking-and-revision step reduces cases where a candidate appears to follow the requested skills or difficulty level but fails to realize the latent reasoning structure specified by the generation target. Implementation details of the checking and revision protocol are provided in the Technical Supplement.

\subsection{Diagnostic Evidence for Realized Difficulty}

A target difficulty specified during synthesis is not sufficient evidence that a generated problem realizes the intended difficulty. A generated problem may appear difficult because it is lengthy, ambiguous, or overcomplicated, rather than because it requires the intended reasoning structure. GaokaoMathGen therefore collects post-generation diagnostic evidence from logical quality, fixed-solver decoding effort, skill-method structural signals, and search-related signals extracted from solver outputs.

For each generated candidate \(\hat{x}_i\), the diagnostic evidence is summarized as follows:
\begin{equation}
r(\hat{x}_i)=
(q_{\mathrm{logic}}(\hat{x}_i),
e_{\mathrm{solve}}(\hat{x}_i),
g_{\mathrm{struct}}(\hat{x}_i),
h_{\mathrm{search}}(\hat{x}_i)),
\label{eq:diagnostic-evidence}
\end{equation}
where \(q_{\mathrm{logic}}\) measures automated logical quality, \(e_{\mathrm{solve}}\) measures fixed-solver decoding effort, \(g_{\mathrm{struct}}\) measures the structural organization of standardized skills and methods, and \(h_{\mathrm{search}}\) captures search-related proxies extracted from the fixed solver's output. These signals are designed to distinguish realized reasoning difficulty from spurious difficulty caused by invalidity, ambiguity, or excessive surface complexity. These four signals are instantiated as follows.

\paragraph{Logical quality filtering.}
Logical quality is evaluated before difficulty analysis, since invalid or ambiguous problems may produce long solver outputs for the wrong reason. We compute logical quality from three dimensions as follows:
\begin{equation}
q_{\mathrm{logic}}(\hat{x}_i)
=
\frac{1}{3}
(q_{\mathrm{clarity}}(\hat{x}_i)
+q_{\mathrm{coherence}}(\hat{x}_i)
+q_{\mathrm{validity}}(\hat{x}_i)),
\label{eq:logic-quality}
\end{equation}
where \(q_{\mathrm{clarity}}\), \(q_{\mathrm{coherence}}\), and \(q_{\mathrm{validity}}\) denote clarity, logical coherence, and mathematical validity, respectively, and each dimension is scored on a 1--5 scale.

A candidate is retained for difficulty analysis only if its average logical-quality score is at least 3 and no individual dimension score is lower than or equal to 2. The retained candidate set is defined as follows:
\begin{equation}
\widehat{\mathcal{X}}_{\mathrm{logic}}
=
\{\hat{x}_i:
q_{\mathrm{logic}}(\hat{x}_i)\ge 3,\;
\min_j q_j(\hat{x}_i)>2\},
\label{eq:logic-filter}
\end{equation}
where \(q_j\) ranges over clarity, logical coherence, and mathematical validity.

\paragraph{Fixed-solver decoding effort.}
For each candidate retained in \(\widehat{\mathcal{X}}_{\mathrm{logic}}\), a fixed reasoning model \(F\) solves the problem under the same prompt and decoding configuration. The fixed-solver decoding effort is defined as follows:
\begin{equation}
e_{\mathrm{solve}}(\hat{x}_i)
=
\mathrm{tokens}(F(\hat{x}_i)),
\qquad
\hat{x}_i\in\widehat{\mathcal{X}}_{\mathrm{logic}},
\label{eq:solver-effort}
\end{equation}
where \(\mathrm{tokens}(\cdot)\) denotes the completion length. This signal provides a consistent proxy for comparing fixed-solver effort across generated candidates, but it should be interpreted as diagnostic evidence rather than as a direct measure of human-perceived difficulty.

\paragraph{Skill-method structural signal.}
To connect difficulty evidence with the structure-level control signals used during generation, GaokaoMathGen computes a topology-only skill-method structural signal. We build an undirected co-occurrence graph over the standardized vocabularies as follows:
\begin{equation}
\mathcal{G}=(\mathcal{V}^S\cup\mathcal{V}^M,E,w),
\label{eq:skill-method-graph}
\end{equation}
where nodes are standardized skills or methods, an edge \((u,v)\in E\) indicates that two labels co-occur in at least one problem, and \(w(u,v)\) records their co-occurrence frequency. The graph uses co-occurrence frequency rather than any prompted difficulty value, so the resulting structural signal remains difficulty-label-free.

For a generated candidate \(\hat{x}_i\), we define its associated skill-method node set as follows:
\begin{equation}
U_i=\mathcal{S}_i\cup\mathcal{M}_i.
\label{eq:skill-method-union}
\end{equation}
The standardized skills and methods associated with the candidate induce a subgraph as follows:
\begin{equation}
\begin{aligned}
\mathcal{G}_i&=\mathcal{G}[U_i]=(U_i,E_i),\\
E_i&=\{(u,v)\in E: u\in U_i,\; v\in U_i\},
\end{aligned}
\label{eq:induced-subgraph}
\end{equation}
where \(\mathcal{G}[U_i]\) denotes the subgraph of \(\mathcal{G}\) induced by the node set \(U_i\). The structural feature vector is computed as follows:
\begin{equation}
g_{\mathrm{struct}}(\hat{x}_i)=\phi_{\mathrm{graph}}(\mathcal{G}_i),
\label{eq:struct-signal}
\end{equation}
where \(\phi_{\mathrm{graph}}\) extracts topology-only features, e.g., node count, edge count, subgraph density, weighted degree centrality, betweenness centrality, PageRank, average co-occurrence frequency, and rare-edge ratio. These features provide label-free evidence about the size, connectivity, centrality, and rarity of the skill-method specification associated with the candidate.

\paragraph{Solver-output search signal.}
GaokaoMathGen further extracts search-related signals from the fixed solver's output. These features are not treated as direct observations of the solver's internal reasoning process. Instead, they provide textual and structural proxies for how the solver organizes its solution. The search-related feature vector is written as follows:
\begin{equation}
h_{\mathrm{search}}(\hat{x}_i)
=
\phi_{\mathrm{search}}(F(\hat{x}_i)),
\label{eq:search-signal}
\end{equation}
where \(\phi_{\mathrm{search}}\) extracts features such as subproblem count, explicit branching patterns, case-analysis count, elimination count, verification count, and an aggregated search-complexity score. The diagnostic evidence is used in the experiments for filtering, ranking, and correlation analysis with fixed-solver decoding effort. High-effort candidates identified by these signals are further sampled for teacher audit, providing a targeted quality check without requiring full manual annotation. Implementation details of the evaluation, solver, and feature-extraction protocol are provided in the Technical Supplement.

\section{Experiments}

We evaluate GaokaoMathGen through three research questions.

\textbf{RQ1 (Structure Extraction).} Does semantic normalization improve cross-annotator comparability of the extracted skills and methods?

\textbf{RQ2 (Controlled Synthesis).} How reliably does synthesis controlled by latent reasoning structures produce logically valid candidates, and do target levels induce distinguishable fixed-solver effort?

\textbf{RQ3 (Difficulty Diagnostics).} Do post-generation signals provide complementary diagnostics of fixed-solver reasoning effort?

We also conduct a seed-matched comparison with an independent reference pipeline and a teacher audit to assess relative performance and residual quality risks. Further results are provided in the Technical Supplement.

\subsection{Experimental Setup}

\paragraph{Data and structure extraction.}
We start with 250 Chinese senior-high-school mathematics problems. Screening removes 33 with missing figures, corrupted content, complete solutions, or material beyond the curriculum, leaving 217 seed problems. Five LLM annotators, i.e., GPT-5.4, DeepSeek-V3.2, Qwen3.5-Plus, Doubao-Seed-2.0-Pro, and Claude Opus 4.6, independently extract latent reasoning structures. Cross-annotator analysis uses the 174 seed problems with valid structured outputs from all five annotators.

\paragraph{Generation and evaluation.}
In the main generation experiment, each LLM is conditioned on a target difficulty 
\(\tau_i\in\{7,8\}\) according to our eight-level difficulty rubric. We instantiate the same synthesis protocol with DeepSeek-V3.2, Doubao-Seed-2.0-Pro, GPT-5.4, and Qwen3.5-Plus. Gemini-2.5-Pro scores each generated statement from 1 to 5 on clarity, logical coherence, and mathematical validity. Following Eq.~\ref{eq:logic-filter}, a candidate is retained when its average score is at least 3 and every dimension is greater than 2. Each retained candidate is then solved by Doubao-Seed-2.0-Pro with the same prompt, high reasoning effort, and a maximum output length of 65,536 tokens. Completion length is used only as a fixed-solver effort signal.

\paragraph{Statistical analysis.}
Our statistical analysis follows the three research questions. For RQ1, we measure cross-annotator overlap using pairwise Jaccard similarity \cite{jaccard1901etude} and examine whether normalization consistently improves this overlap across seeds using a paired Wilcoxon signed-rank test \cite{wilcoxon1945individual}. For RQ2, we test whether target-8 candidates induce greater fixed-solver decoding effort than target-7 candidates within each generator using a one-sided Mann--Whitney \(U\) test \cite{mann1947test}. For RQ3, we use Spearman's rank correlation \cite{spearman1987proof} to quantify the monotonic association between each diagnostic signal and fixed-solver decoding effort. The external comparison uses a paired Wilcoxon signed-rank test, as the candidates from the two pipelines are matched by their source seeds. Holm correction \cite{holm1979simple} is applied separately to the model-wise target-level and external comparisons. Additional statistical details, model configurations, and prompt protocols are provided in the Technical Supplement.

\subsection{Reliability of Structure Extraction}

We first examine whether normalization reduces annotator-specific variation in the skill and method components of the latent reasoning structure. For each seed, we form all \(\binom{5}{2}=10\) unordered pairs of the five annotators. For each pair, we compute the Jaccard similarity and an exact-match indicator for the skill and method sets, both before and after applying the normalization mappings used to construct the generation inputs. We first average each metric over the ten annotator pairs for each seed and then report the mean across the 174 seeds. Table~\ref{tab:normalization-agreement} summarizes the resulting cross-annotator agreement.

\begin{table}[t]
\centering
\small
\setlength{\tabcolsep}{5pt}
\caption{Cross-annotator agreement before and after semantic normalization. Exact denotes exact-set agreement.}
\label{tab:normalization-agreement}
\begin{tabular}{llrr}
\hline
Label & Metric & Raw & Normalized \\
\hline
Skills  & Jaccard & 0.015 & \textbf{0.261} \\
Skills  & Exact & 0.001 & \textbf{0.023} \\
Methods & Jaccard & 0.045 & \textbf{0.211} \\
Methods & Exact & 0.006 & \textbf{0.044} \\
\hline
\end{tabular}
\end{table}

Normalization raises mean pairwise Jaccard similarity from 0.015 to 0.261 for skills and from 0.045 to 0.211 for methods. Both improvements are significant under paired seed-level Wilcoxon tests ($p<10^{-29}$). Exact-set agreement also increases, although it remains low, as annotators may select complementary labels for the same problem. The numerical difficulty annotations show stronger agreement, with mean pairwise Spearman correlations of 0.801 for abstraction and 0.774 for computation. These results support the use of normalized annotations as more comparable structure-level control signals, without assuming that normalization removes all annotation uncertainty.

\subsection{Generation Quality and Target-Level Separation}

Starting from the 174 common seeds and four generators, the 696 possible seed--generator instances are reduced to 663 generation inputs after excluding 33 annotations without usable mathematical kernels, of which 609 yield candidate problems with complete logical-quality scores. Table~\ref{tab:target-control} summarizes logical-quality retention and fixed-solver effort. Overall, 475 of 609 scored candidates pass the logical-quality filter. GPT-5.4 has the highest retention rate at 94.1\%, followed by Doubao-Seed-2.0-Pro at 85.1\%. The variation across generators shows that a shared latent reasoning structure and synthesis protocol do not guarantee equally reliable realization by different LLMs.

\begin{table}[t]
\centering
\small
\setlength{\tabcolsep}{3.4pt}
\caption{Logical-quality retention and mean fixed-solver effort at target levels 7 and 8. $E_7$ and $E_8$ are computed over retained candidates.}
\label{tab:target-control}
\begin{tabular}{lrrrr}
\hline
Generator & Pass (\%) & $E_7$ & $E_8$ & $\Delta$ (\%) \\
\hline
DeepSeek-V3.2 & 59.5 & 15,083 & 20,198 & +33.9 \\
Doubao-Seed-2.0-Pro   & 85.1 & 18,717 & 22,927 & +22.5 \\
GPT-5.4  & \textbf{94.1} & 21,078 & 24,041 & +14.1 \\
Qwen3.5-Plus     & 68.3 & 18,611 & 23,302 & +25.2 \\
\hline
All      & 78.0 & 18,909 & 22,996 & +21.6 \\
\hline
\end{tabular}
\end{table}

Among retained candidates, target-8 problems require 22,996 tokens on average, compared with 18,909 tokens for target 7, an increase of 21.6\%. The association is positive for all four generators, and every within-generator comparison remains significant after Holm correction ($p<0.05$). A regression with generator fixed effects estimates an adjusted target-8 increase of 4,040 tokens ($p<10^{-5}$). The aggregate logical retention rates are 80.6\% for target 7 and 75.8\% for target 8; this difference is not statistically significant under Fisher's exact test ($p=0.170$). Thus, the two requested levels are separated in fixed-solver effort without a significant aggregate change in logical retention. This result concerns controlled solver effort and does not by itself establish equivalent separation in human difficulty.

The independent structural checker is also necessary in practice. On development batches, it requests revision for 98.3\% of candidates produced by the initial synthesis prompt and 85.1\% after prompt refinement. After one checking-and-revision round, the corresponding no-revision rates are 0.8\% and 11.9\%. These development results are not a controlled ablation, but they show that the generator's own blueprint is insufficient evidence that the intended structural requirements are satisfied. Detailed checker outcomes are reported in the Technical Supplement.

\subsection{Diagnostic Evidence for Realized Difficulty}

Table~\ref{tab:diagnostic-correlation} evaluates the topology-only skill--method signal and solver-output search signals against fixed-solver completion length on the 475 retained candidates. Records are matched by both generator and seed identifier to avoid collisions between candidates generated from the same seed.

\begin{table}[t]
\centering
\small
\setlength{\tabcolsep}{5pt}
\caption{Spearman correlation with fixed-solver completion length on the retained high-difficulty batch ($N=475$).}
\label{tab:diagnostic-correlation}
\begin{tabular}{llrr}
\hline
Evidence & Feature & $\rho$ & $p$ \\
\hline
Graph & Node count & 0.067 & 0.146 \\
Graph & Edge count & 0.053 & 0.246 \\
Solver output & Branching factor & 0.438 & $<0.001$ \\
Solver output & Case analysis & 0.490 & $<0.001$ \\
Solver output & Verification & 0.322 & $<0.001$ \\
Solver output & Search score & \textbf{0.539} & $<0.001$ \\
\hline
\end{tabular}
\end{table}

The topology-only features are weak in this range-restricted 7--8 batch. In contrast, the search-related features have moderate positive associations with completion length, and the aggregate search score reaches $\rho=0.539$. The same trend holds within every generator, with correlations from 0.462 to 0.636. These signals characterize how the fixed solver organizes its output, but they are not independent human-difficulty measurements, as both the features and completion length are derived from the same solution. We therefore treat them as complementary diagnostics rather than as ground-truth difficulty labels.

\subsection{External Evidence and Error Analysis}

We compare GaokaoMathGen with an independently developed industrial reference pipeline on seed-matched candidates. Candidates from the two pipelines are aligned by source-seed identifier and solved by the same fixed solver. Since the pipelines differ in their generation procedures and post-generation selection, this comparison serves as an external reference rather than a controlled component ablation. Table~\ref{tab:reference-comparison} reports the resulting fixed-solver effort.

\begin{table}[t]
\centering
\small
\setlength{\tabcolsep}{2.5pt}
\caption{Seed-matched comparison of mean fixed-solver completion length. Ref. denotes the industrial reference pipeline, and $\Delta$ denotes the relative difference (\%). $^{*}$ indicates Holm-corrected $p<0.05$.}
\label{tab:reference-comparison}
\begin{tabular}{@{}lrrrr@{}}
\hline
Generator & $N$ & Ref. & Ours & $\Delta$ \\
\hline
Doubao-Seed-2.0-Pro   & 131 & 20,245 & 21,031 & +3.9 \\
Gemini-3.1-Pro   & 153 & 19,398 & 19,296 & $-0.5$ \\
Qwen3.5-Plus     & 112 & 19,286 & 21,082 & +9.3 \\
GPT-5.2  & 146 & 18,198 & 19,965 & +9.7 \\
GPT-5.4  & 156 & 20,447 & 22,686 & +11.0$^{*}$ \\
\hline
\end{tabular}
\end{table}

GaokaoMathGen produces higher mean fixed-solver completion lengths in four of the five matched subsets, with a macro-average relative difference of $+6.7\%$. After Holm correction, only the GPT-5.4 comparison remains significant ($p=0.0145$). These results support comparable fixed-solver effort under matched seeds, with a significant increase in the GPT-5.4 subset, rather than uniform superiority over the reference pipeline.

Finally, a high-school mathematics teacher audits the 60 GPT-5.4 candidates with the largest fixed-solver completion lengths from both difficulty and quality perspectives. The teacher judges all 60 candidates to reach the difficulty expected of a Gaokao finale or penultimate problem. Regarding quality, 59 of the 60 candidates (98.3\%) contain no severe logical issue. Minor concerns in 21 cases mainly involve curriculum boundaries or nonstandard notation and do not invalidate the problems. The only severe case (1.7\%) contains an ambiguous dependency between subproblems. These results provide targeted human evidence that candidates ranked highest by fixed-solver effort fall within the intended high-difficulty range, while severe logical failures are rare in this audited subset. Since the candidates are deliberately selected by fixed-solver effort, the audit does not estimate corpus-wide quality rates or establish human difficulty separation between target levels 7 and 8.

\section{Conclusion and Future Work}

We introduced GaokaoMathGen, to our knowledge the first framework dedicated to generating Chinese Gaokao mathematics finale problems. It converts seed problems into standardized latent reasoning structures, whose core identities and pivot steps guide constrained synthesis. Experiments demonstrate the value of semantic normalization, structure-guided generation, and post-generation diagnostics for high-difficulty candidates. After validation, its candidates can support teacher-assisted question development, advanced student practice, and training and evaluation data for mathematical reasoning models. Future work will incorporate broader expert feedback and multi-solver evidence, directly evaluate structural realization, and extend the framework to multimodal mathematics problems.

\clearpage
\appendix
\setcounter{secnumdepth}{2}
\setcounter{table}{0}
\setcounter{equation}{0}
\renewcommand{\thetable}{S\arabic{table}}
\renewcommand{\theequation}{S\arabic{equation}}
\renewcommand{\theHtable}{supp.\arabic{table}}
\renewcommand{\theHequation}{supp.\arabic{equation}}
\suppressfloats[t]
\section*{Technical Supplement}
\pdfbookmark[1]{Technical Supplement}{supplement}
This supplement reports additional experimental results and describes the prompting and evaluation protocols used in the main paper.

\section{Additional Experimental Details}
\label{app:experiments}

Unless otherwise stated, all solver-effort analyses use Doubao-Seed-2.0-Pro under a fixed prompt, high reasoning effort, and a maximum output length of 65,536 tokens.

\subsection{Data Processing and Coverage}

Table~\ref{tab:app-data-flow} summarizes the construction of the seed set. The common 174-seed subset contains 37 function-and-derivative problems, 40 analytic-geometry problems, 31 sequence problems, 42 probability-and-statistics problems, and 24 solid-geometry problems.

\begin{table}[!htbp]
\centering
\small
\caption{Seed-data processing.}
\label{tab:app-data-flow}
\begin{tabular}{lrr}
\hline
Stage & Problems & Retained (\%) \\
\hline
Raw collection & 250 & 100.0 \\
After automatic screening & 217 & 86.8 \\
Successful for all five annotators & 174 & 69.6 \\
\hline
\end{tabular}
\end{table}

The five annotators produce 1,039 valid structured outputs out of 1,085 attempted annotations, giving an overall extraction success rate of 95.8\%. Table~\ref{tab:app-annotation-coverage} gives the per-model coverage. The common subset is used for agreement analyses so that every seed has the same ten annotator pairs.

\begin{table}[!htbp]
\centering
\small
\setlength{\tabcolsep}{5pt}
\caption{Structured-annotation success rates.}
\label{tab:app-annotation-coverage}
\begin{tabular}{lrr}
\hline
Annotator & Valid/Attempted & Success (\%) \\
\hline
GPT-5.4 & 217/217 & 100.0 \\
Qwen3.5-Plus & 217/217 & 100.0 \\
DeepSeek-V3.2 & 216/217 & 99.5 \\
Doubao-Seed-2.0-Pro & 215/217 & 99.1 \\
Claude Opus 4.6 & 174/217 & 80.2 \\
\hline
All & 1,039/1,085 & 95.8 \\
\hline
\end{tabular}
\end{table}

Table~\ref{tab:app-candidate-flow} summarizes candidate flow in the main generation experiment. Starting from the 174 common seeds and four generators, the experiment contains 696 possible seed--generator instances. Excluding 33 annotations without usable mathematical kernels leaves 663 synthesis inputs. Of these, 609 produce candidate problems with complete logical-quality scores, and 475 pass the logical-quality filter.

\begin{table}[!htbp]
\centering
\small
\caption{Candidate flow in the main generation experiment.}
\label{tab:app-candidate-flow}
\begin{tabular}{lr}
\hline
Stage & Instances \\
\hline
Possible seed--generator instances & 696 \\
Usable synthesis inputs & 663 \\
Candidates with complete quality scores & 609 \\
Candidates retained for difficulty analysis & 475 \\
\hline
\end{tabular}
\end{table}

Table~\ref{tab:app-model-config} summarizes the recorded model-side configurations. We retain provider defaults when a stored request does not specify an additional decoding parameter.

\begin{table*}[t]
\centering
\small
\setlength{\tabcolsep}{5pt}
\caption{Recorded model configurations used in the main pipeline.}
\label{tab:app-model-config}
\begin{tabular}{lll}
\hline
Role & Model & Recorded configuration \\
\hline
Structure annotation & DeepSeek-V3.2 & High reasoning effort; 65,536 output tokens \\
Structure annotation & Doubao-Seed-2.0-Pro & High reasoning effort; 65,536 output tokens \\
Structure annotation & GPT-5.4 & Extra-high reasoning effort; detailed summary; 65,536 output tokens \\
Structure annotation & Qwen3.5-Plus & Thinking enabled; 65,536 output tokens \\
Structure annotation & Claude Opus 4.6 & 65,536 output tokens \\
Main synthesis & DeepSeek-V3.2 & High reasoning effort; 65,536 output tokens \\
Main synthesis & Doubao-Seed-2.0-Pro & High reasoning effort; 65,536 output tokens \\
Main synthesis & GPT-5.4 & Extra-high reasoning effort; detailed summary; 65,536 output tokens \\
Main synthesis & Qwen3.5-Plus & Provider-default recorded request \\
Logical evaluator & Gemini-2.5-Pro & Fixed three-dimension scoring prompt \\
Fixed solver & Doubao-Seed-2.0-Pro & High reasoning effort; 65,536 output tokens \\
\hline
\end{tabular}
\end{table*}

The semantic-merging stage starts from 2,716 globally unique skill strings. Subject-conditioned partitioning creates 2,768 skill entries because the same surface string may occur in more than one domain; these entries are grouped into 238 canonical clusters. The corresponding method inventory contains 1,379 globally unique strings grouped into 46 canonical clusters. We retain unmatched labels to preserve annotation coverage, and the synthesis experiments use the normalized labels written to the generation inputs.

\subsection{Structure-Extraction Agreement}

For label sets $A$ and $B$, Jaccard similarity is computed as follows:
\begin{equation}
J(A,B)=\frac{|A\cap B|}{|A\cup B|}.
\label{eq:app-jaccard}
\end{equation}
For each seed and label type, we average over the ten pairs of five annotators. Raw hierarchical skill paths are reduced to their leaf labels before comparison, matching the preprocessing used by the normalization pipeline. Significance is assessed with a one-sided paired Wilcoxon signed-rank test over the 174 seed-level averages, with the alternative that normalization increases agreement. Skills improve from 0.0153 to 0.2610 ($p=1.32\times10^{-30}$), and methods improve from 0.0453 to 0.2113 ($p=1.27\times10^{-29}$). Exact-set agreement changes from 0.0006 to 0.0230 for skills and from 0.0063 to 0.0443 for methods.

For methods, the reported $p$-value is slightly above the $10^{-29}$ bound stated in the main text.

Table~\ref{tab:app-difficulty-agreement} reports agreement for the two numerical difficulty annotations. Pairwise correlations are computed across seeds for each annotator pair and then averaged. Within-seed standard deviation measures dispersion among the five annotators.

\begin{table}[!htbp]
\centering
\small
\caption{Agreement of numerical seed-difficulty annotations.}
\label{tab:app-difficulty-agreement}
\begin{tabular}{lrr}
\hline
Dimension & Pairwise $\rho$ & Within-seed SD \\
\hline
Abstraction & 0.801 & 0.670 \\
Computation & 0.774 & 0.803 \\
\hline
\end{tabular}
\end{table}

\subsection{Structural Checker Development}
\label{app:checker-analysis}

Table~\ref{tab:app-checker} details the development-batch outcomes reported in the main text. A nonempty \texttt{ques\_revised} block indicates a revision, an empty block indicates no revision, and outputs without a matching block are counted separately. Percentages use all checker outputs for the corresponding prompt and round, including outputs without a matching block. In the second round, the revised statement is checked when available, and the original statement is reused otherwise.

\begin{table*}[t]
\centering
\small
\setlength{\tabcolsep}{5pt}
\caption{Structural-checker outcomes on the two development batches. Values in parentheses are percentages of $N$.}
\label{tab:app-checker}
\begin{tabular}{llrrrr}
\hline
Synthesis prompt & Checker round & $N$ & Revision & No revision & Missing block \\
\hline
Initial & First & 663 & 652 (98.3) & 4 (0.6) & 7 \\
Initial & Second & 663 & 650 (98.0) & 5 (0.8) & 8 \\
Refined & First & 631 & 537 (85.1) & 46 (7.3) & 48 \\
Refined & Second & 631 & 546 (86.5) & 75 (11.9) & 10 \\
\hline
\end{tabular}
\end{table*}

The two batches differ in both synthesis instructions and candidate counts. These are development results. The logical-quality retention and solver-effort results in Table~\ref{tab:target-control} use the initial statements from the main generation batch.

\subsection{Detailed Target-Level Analysis}

Target difficulty values $\tau_i\in\{7,8\}$ are assigned with random seed 42 and are not allowed to fall below the annotator's abstraction score for the seed. The corresponding implementation field is named \texttt{target\_abstraction\_score}; throughout the paper, it is denoted by $\tau_i$ because it serves as the requested difficulty level during synthesis. Table~\ref{tab:app-target-level} gives target-specific logical retention and solver effort. The tests compare retained target-8 and target-7 candidates separately for each generator. All four one-sided Mann--Whitney tests remain significant after Holm correction.

\begin{table*}[t]
\centering
\small
\setlength{\tabcolsep}{5pt}
\caption{Detailed target-level results. Effort is the mean completion length over retained candidates.}
\label{tab:app-target-level}
\begin{tabular}{lrrrrrr}
\hline
Generator & Target-7 pass & Target-8 pass & $E_7$ ($N$) & $E_8$ ($N$) & $\Delta$ (\%) & Holm $p$ \\
\hline
DeepSeek-V3.2 & 36/66 (54.5) & 36/55 (65.5) & 15,083 (36) & 20,198 (36) & +33.9 & 0.0133 \\
Doubao-Seed-2.0-Pro & 59/67 (88.1) & 72/87 (82.8) & 18,717 (59) & 22,927 (72) & +22.5 & 0.0044 \\
GPT-5.4 & 76/76 (100.0) & 84/94 (89.4) & 21,078 (76) & 24,041 (84) & +14.1 & 0.0144 \\
Qwen3.5-Plus & 53/69 (76.8) & 59/95 (62.1) & 18,611 (53) & 23,302 (59) & +25.2 & 0.0280 \\
\hline
All & 224/278 (80.6) & 251/331 (75.8) & 18,909 (224) & 22,996 (251) & +21.6 & -- \\
\hline
\end{tabular}
\end{table*}

To account for differences in average solver effort across generators, we conduct a robustness analysis using the following ordinary least-squares model:
\begin{equation}
e_{\mathrm{solve},i}
=
\beta_0+\beta_1\mathbf{1}(\tau_i=8)
+\sum_{g=1}^{3}\alpha_g\mathbf{1}(G_i=g)+\epsilon_i,
\label{eq:app-effort-regression}
\end{equation}
where $G_i$ denotes the candidate's generator and one of the four generators is the reference category. Fitted on the 475 retained candidates, the model gives $\hat{\beta}_1=4{,}040.3$ tokens. An HC1 heteroscedasticity-consistent standard error of 840.4 yields a two-sided $t$-test result of $p=2.06\times10^{-6}$. Thus, the reported 4,040-token difference is the target-8 coefficient adjusted for generator-level differences, rather than the unadjusted difference between the two target-level means. Aggregate logical retention is 80.6\% at target 7 and 75.8\% at target 8; Fisher's exact test finds no significant difference ($p=0.170$). These results confirm target-level separation in fixed-solver effort after accounting for generator variation.

\subsection{Cross-Domain Results}

Table~\ref{tab:app-subject-results} reports results by the five source domains. Retention ranges from 70.0\% in solid geometry to 83.1\% in probability and statistics. Analytic geometry induces the largest mean fixed-solver effort. These results indicate that the pipeline is not confined to a single domain, while also revealing domain-dependent reliability.

\begin{table}[!htbp]
\centering
\small
\setlength{\tabcolsep}{4pt}
\caption{Logical retention and fixed-solver effort by source domain.}
\label{tab:app-subject-results}
\begin{tabular}{lrrr}
\hline
Domain & Retained/All & Pass (\%) & Mean effort \\
\hline
Functions/derivatives & 101/128 & 78.9 & 21,386 \\
Analytic geometry & 110/145 & 75.9 & 24,245 \\
Sequences & 85/108 & 78.7 & 18,693 \\
Probability/statistics & 123/148 & 83.1 & 19,855 \\
Solid geometry & 56/80 & 70.0 & 20,529 \\
\hline
\end{tabular}
\end{table}

\subsection{Additional Diagnostic Analyses}

\paragraph{Within-generator stability.}
For the high-difficulty batch, search features are recomputed from each generator-specific solver-output file and joined with graph records using the composite key consisting of generator and seed identifier. This prevents candidates derived from the same seed by different generators from being conflated. Table~\ref{tab:app-search-by-generator} reports the association between the aggregate search score and completion length within each generator. The correlation remains significant in all four subsets, reducing the possibility that the aggregate result is driven only by between-generator differences.

\begin{table}[!htbp]
\centering
\small
\caption{Search-score correlations within each generator.}
\label{tab:app-search-by-generator}
\begin{tabular}{lrr}
\hline
Generator & $N$ & Spearman $\rho$ \\
\hline
DeepSeek-V3.2 & 72 & 0.636 \\
Doubao-Seed-2.0-Pro & 131 & 0.528 \\
GPT-5.4 & 160 & 0.554 \\
Qwen3.5-Plus & 112 & 0.462 \\
\hline
\end{tabular}
\end{table}

\paragraph{Text-only baseline.}
We compare solver-search features with simple lexical indicators over a broader set of 3,160 solver outputs. The aggregate text score reaches only $\rho=0.236$, compared with $\rho=0.578$ for the aggregate search score on the same broader corpus. Table~\ref{tab:app-text-baseline} lists the strongest lexical features. This comparison shows that the search features capture aspects of solution organization beyond surface lexical cues.

\begin{table}[!htbp]
\centering
\small
\caption{Text-only correlations with completion length ($N=3{,}160$).}
\label{tab:app-text-baseline}
\begin{tabular}{lr}
\hline
Feature & Spearman $\rho$ \\
\hline
Aggregate text score & 0.236 \\
Inverse-problem indicator & 0.151 \\
Quantifier depth & 0.143 \\
Existence indicator & 0.135 \\
Condition indicator & 0.132 \\
Parameter indicator & 0.129 \\
\hline
\end{tabular}
\end{table}

\subsection{Seed-Matched External Comparison}
\label{app:external-comparison}

We compare GaokaoMathGen with an independently developed industrial generation pipeline. Candidates are aligned by source-seed identifier, and both sides are solved with the same fixed solver. We additionally instantiate the same structure-guided synthesis protocol with Gemini-3.1-Pro and GPT-5.2 to broaden model coverage in this comparison. Table~\ref{tab:app-reference-comparison} reports means, medians, and paired Wilcoxon results, providing an external benchmark under seed-matched evaluation.

\begin{table*}[t]
\centering
\small
\setlength{\tabcolsep}{5pt}
\caption{Seed-matched fixed-solver effort. $^{*}$ denotes Holm-corrected $p<0.05$.}
\label{tab:app-reference-comparison}
\begin{tabular}{lrrrrrr}
\hline
Generator & $N$ & Ref. mean & Ours mean & Ref. median & Ours median & Holm $p$ \\
\hline
Doubao-Seed-2.0-Pro & 131 & 20,245 & 21,031 & 18,555 & 19,327 & 0.9596 \\
Gemini-3.1-Pro & 153 & 19,398 & 19,296 & 18,975 & 18,483 & 0.9596 \\
Qwen3.5-Plus & 112 & 19,286 & 21,082 & 17,675 & 18,540 & 0.9596 \\
GPT-5.2 & 146 & 18,198 & 19,965 & 17,915 & 18,521 & 0.0857 \\
GPT-5.4 & 156 & 20,447 & 22,686 & 19,834 & 21,480 & 0.0145$^{*}$ \\
\hline
\end{tabular}
\end{table*}

The mean relative differences are $+3.9\%$, $-0.5\%$, $+9.3\%$, $+9.7\%$, and $+11.0\%$, respectively, with a macro average of $+6.7\%$. After Holm correction, the GPT-5.4 comparison remains significant. Overall, GaokaoMathGen achieves competitive or higher fixed-solver effort across the seed-matched subsets.

\subsection{Teacher Audit}

For the teacher audit, all 160 retained GPT-5.4 candidates are re-solved in a dedicated ranking run using the same solver model and decoding settings as the main analysis. The ranking run uses a separate fixed instruction, and its completion lengths are used only to select the 60 highest-effort candidates for the audit. A high-school mathematics teacher assesses the difficulty and quality of every selected problem individually. All 60 are judged to reach the difficulty expected of a Gaokao finale or penultimate problem, and 59 contain no severe logical issue. Table~\ref{tab:app-teacher-audit} summarizes these item-level judgments, providing targeted human evidence for the quality and difficulty of high-effort candidates.

\begin{table}[!htbp]
\centering
\small
\caption{Teacher audit of the 60 GPT-5.4 candidates selected by the dedicated ranking run.}
\label{tab:app-teacher-audit}
\begin{tabular}{lrr}
\hline
Assessment outcome & Count & Percentage \\
\hline
Finale/penultimate difficulty & 60 & 100.0 \\
No severe logical issue & 59 & 98.3 \\
Severe logical issue & 1 & 1.7 \\
\hline
\end{tabular}
\end{table}

The 59 candidates without a severe logical issue include the 21 candidates with minor concerns noted in the main text. The single severe case concerns an ambiguous dependency between subproblems.

\section{Prompt and Protocol Details}
\label{app:prompts}

\paragraph{Structure annotation.}
Each annotator receives the seed statement and returns problem type, subject area, skills, methods, core identity, pivot step, abstraction and computation scores, a difficulty anchor, and auxiliary structural fields. Skills are expressed as curriculum knowledge components, while methods describe reusable solution strategies. The output is parsed into a fixed JSON schema. Skill normalization is performed separately within the five source domains, whereas method normalization is performed globally.

\paragraph{Semantic normalization.}
We group raw leaf labels by semantic equivalence using multiple LLM proposals and adopt a mapping when at least two proposals agree. Before mapping, we remove hierarchy prefixes and parenthetical elaborations. We retain unmatched labels to preserve information, and the agreement experiment uses the resulting normalized labels.

\paragraph{Structure-guided synthesis.}
The synthesis prompt provides skills, methods, the reference core identity, the reference pivot step, target difficulty, and difficulty anchor. In the stored prompt, target difficulty is represented by the field \texttt{target\_abstraction\_score}. The generator returns a construction blueprint, a complete problem statement inside a dedicated tag, and a reference solution. The prompt requires kernel grounding, concealment by concept wrapping or structural reorganization, shortcut analysis, condition-redundancy checks, and subproblem progression. It supports domain-consistent adaptation of the reference kernel to integrate the specified skills and methods, while requiring the final core identity and pivot step to be recorded explicitly in the blueprint.

\paragraph{Structural checking and revision.}
The checker receives the generated statement together with the final core identity and pivot step, but not the generator's full self-explanation. The following is an English rendering of the final checker prompt. The operational prompt is written in Chinese.

\begin{quote}
\small
\textbf{Role.} Act as a logic reviewer for Chinese senior-high-school mathematics problem design and identify defects rather than evaluating stylistic appeal.

\textbf{Inputs.} The final \texttt{core\_identity}, the final \texttt{pivot\_step}, and the \texttt{generated\_problem}.

\textbf{Task.} Determine whether the problem realizes the supplied mathematical kernel and whether its statement contains structural or mathematical defects.

\textbf{Checks.}
(1) Trace each condition to the core identity, pivot step, and conclusion.
(2) Test whether special-value substitution, direct expansion, numerical trial, or another elementary route bypasses the pivot step.
(3) Remove each condition in turn and identify conditions that do not affect unique solvability.
(4) For multi-part problems, examine whether later subproblems meaningfully build on earlier ones.
(5) Check variable domains, consistency of conditions, leakage between subproblems, boundary cases, notation, and solvability.

\textbf{Output.} Return a check-by-check report, a severity-ordered defect list, and, when a defect is found, a revised statement enclosed by the designated \texttt{ques\_revised} tag together with a concise revision note.
\end{quote}

When a violation is identified, the returned statement is checked once more using the same final protocol.

\paragraph{Logical-quality evaluator.}
The evaluator scores clarity and grammar, logical coherence and completeness, and mathematical validity and solvability from 1 to 5. The prompt requests a short justification followed by a machine-readable score list. A problem is retained only when the mean is at least 3 and no score is at most 2.

\paragraph{Fixed solver.}
Every retained problem is prefixed with the same instruction requesting a detailed derivation and final answer. The solver uses high reasoning effort and a 65,536-token maximum. We record completion tokens and retain the generated solution for deterministic feature extraction.

\paragraph{Search-feature extraction.}
No additional LLM prompt is used for search features. A deterministic parser counts explicit subproblems, case-analysis constructions, elimination expressions, verification expressions, and explicit branch markers in the solver output. The aggregate search score combines branching, case-analysis, and verification signals after normalization. This distinction is important because the extracted features describe observable output organization rather than hidden internal model states.

\paragraph{Teacher audit.}
The ranking run uses Doubao-Seed-2.0-Pro with high reasoning effort, a 65,536-token maximum, and a fixed Chinese instruction requesting a complete solution and final answer. Its completion lengths are used to rank the GPT-5.4 audit pool. For each selected problem, the teacher judges whether its difficulty reaches the level of a Gaokao finale or penultimate problem and checks for severe logical defects.

\end{document}